\def\year{2021}\relax
\documentclass[letterpaper]{article} 
\usepackage{aaai21}  
\usepackage{times}  
\usepackage{helvet} 
\usepackage{courier}  
\usepackage[hyphens]{url}  
\usepackage{graphicx} 
\usepackage{graphicx}  
\usepackage{natbib}  
\usepackage{caption} 
\usepackage{glossaries}
\newacronym{PDDL}{PDDL}{Planning Domain Definition Language}
\newacronym[plural=STN,firstplural=Simple Temporal Networks (STN)]{STN}{STN}{Simple Temporal Network}
\newacronym{TFD}{TFD}{Temporal Fast Downward}
\newacronym{STP}{STP}{Simple Temporal Problem}
\newacronym{SMT}{SMT}{Satisfiability Modulo Theories}
\newacronym{AI}{AI}{Artificial Intelligence}
\newacronym{MIP}{LP}{Linear Programming}
\newacronym{SAT}{SAT}{Satisfiability Problem}
\newacronym{SBB}{SBB}{Spatial Branch-and-Bound}
\newacronym{MINLP}{MINLP}{Mixed Integer Non-Linear Programming}
\newacronym{CP}{CP}{Constraint Programming}
\newacronym{UAV}{UAV}{Unmanned Aerial Vehicle}
\newacronym{AUV}{AUV}{Autonomous Underwater Vehicle}
\newacronym{MAV}{MAV}{Micro Aerial Vehicles}
\newacronym{QA}{QA}{Quality Assurance}

\newglossaryentry{colin}
{name={OPTIC},
	description={The planner we modify in this work}
}

\newglossaryentry{opticii}{
	name={\mbox{OPTIC-II}},
	description={The new planner}
}

\newglossaryentry{I}
{
	name = {\ensuremath{I}},
    sort = {I},
    description = {initial state}
}

\newglossaryentry{G}
{
	name = {\ensuremath{G}},
    sort = {G},
    description = {Goal state}
}

\newglossaryentry{M}
{
	name = {\ensuremath{F}},
    sort = {M},
    description = {Metric optimization problem}
}

\newglossaryentry{P}
{
	name = {\ensuremath{\mathbf{P}}},
    sort = {P},
    description = {Set of propositions}
}

\newglossaryentry{V}
{
	name = {\ensuremath{\mathbf{V}}},
    sort = {V},
    description = {Set of numerical Variables}
}

\newglossaryentry{A}
{
	name = {\ensuremath{\mathbf{A}}},
    sort = {A},
    description = {set of available actions}
}

\newglossaryentry{pre}
{
	name = {\ensuremath{pre}},
    sort = {pre},
    description = {precondition}
}

\newglossaryentry{pres}
{
	name = {\ensuremath{pre_{\vdash}}},
    sort = {pres},
    description = {start condition}
}

\newglossaryentry{pree}
{
	name = {\ensuremath{pre_{\dashv}}},
    sort = {pree},
    description = {end condition}
}

\newglossaryentry{preinv}
{
	name = {\ensuremath{pre_{\leftrightarrow}}},
    sort = {preinv},
    description = {invariant condition}
}

\newglossaryentry{eff}
{
	name = {\ensuremath{\mathit{{eff}}}},
    sort = {eff},
    description = {effect}
}

\newglossaryentry{effs}
{
	name = {\ensuremath{\gls{eff}_{\vdash}}},
    sort = {effs},
    description = {effect at start}
}

\newglossaryentry{effe}
{
	name = {\ensuremath{\gls{eff}_{\dashv}}},
    sort = {effe},
    description = {effect at end}
}

\newglossaryentry{effadd}
{
	name = {\ensuremath{\gls{eff}_{+}}},
	sort = {effadd},
	description = {Effect that adds a preposition}
}

\newglossaryentry{effdelete}
{
	name = {\ensuremath{\gls{eff}_{-}}},
	sort = {effdelete},
	description = {Effect that deletes a preposition}
}

\newglossaryentry{effnum}
{
	name = {\ensuremath{\gls{eff}_{num}}},
	sort = {effdelete},
	description = {Effect that changes a numerical value}
}

\newglossaryentry{effinv}
{
	name = {\ensuremath{\gls{eff}_{\leftrightarrow}}},
    sort = {effinv},
    description = {effect over all the duration of the action}
}

\newglossaryentry{dur}
{
	name = {\ensuremath{d}},
    sort = {d},
    description = {effect}
}

\newglossaryentry{ti}
{
	name = {\ensuremath{t_i}},
    sort = {ti},
    description = {Timestamp of step \ensuremath{i}}
}

\newglossaryentry{vi}
{
	name = {\ensuremath{v_i}},
    sort = {vi},
    description = {Value of \ensuremath{v} before step \ensuremath{i}}
}

\newglossaryentry{v}
{
	name = {\ensuremath{v}},
	sort = {v},
	first = {\ensuremath{v\in\gls{V}}},
	description = {Numerical variable \ensuremath{v \in V} }
}

\newglossaryentry{dvi}
{
	name = {\ensuremath{\delta v_i}},
    sort = {dvi},
    description = {Rate of change of \ensuremath{v} after step \ensuremath{i}}
}

\newglossaryentry{tprev}{
	name={\ensuremath{t_{prev}}},
	description={Time in which $v$ was last computed}
}

\newglossaryentry{vprev}{
	name={\ensuremath{v_{prev}}},
	description={Value of $v$ at the last time it was computed}
}

\newglossaryentry{tpe}{
	name={\ensuremath{t_{p_{eff}}}},
	description={Time in which $v$ had an effect start/end}
}

\newglossaryentry{vpe}{
	name={\ensuremath{v_{p_{eff}}}},
	description={Value of $v$ had an effect start/end}
}
\usepackage{tikz}
\usetikzlibrary{patterns,positioning,fit}
\usepackage{tikzscale}
\usepackage{amsmath}
\usepackage{amsthm}
\usepackage{dsfont}
\usepackage{subcaption}
\usepackage{multirow}
\usepackage{fontawesome}
\usepackage{hhline}

\title{Improving Search by Utilizing State Information in OPTIC Planners Compilation to LP}

\author{Elad Denenberg\textsuperscript{\rm 1}, Amanda Coles\textsuperscript{\rm 2}, and Derek Long\textsuperscript{\rm 2}\\
	\textsuperscript{\rm 1}University of Haifa, Abba Khoushy Ave 199, Haifa, Israel\\
	\textsuperscript{\rm 2}King's College London, 30 Aldwych, London, United Kingdom}

\LetLtxMacro{\originaleqref}{\eqref}
\renewcommand{\eqref}{Eq.~\originaleqref}
\newcommand\figref{Fig.~\ref}

\newcommand{\tabref}{Table~\ref}
\newcommand{\secref}{Section~\ref}

\usepackage[draft]{changes} 
\usepackage[]{todonotes}  
\definechangesauthor[name={Elad}, color=blue]{ed}
\definechangesauthor[name={Amanda}, color=red]{ac} 
\definechangesauthor[name={Derek}, color =purple]{dl}
\definechangesauthor[name={Reviewer},color=orange]{Rev}

\makeatletter
\let\Changes@Markup@Deleted\@gobble %
\tikzset{nomorepostaction/.code=\let\tikz@postactions\pgfutil@empty}
\makeatother

\makeatletter
\def\nocopyright{\gdef\copyright@on{}}
\makeatother

\nocopyright

\begin{document}

\maketitle

\begin{abstract}
  Automated planners are computer tools that allow autonomous agents to make strategies and decisions by determining a set of actions for the agent that to take, which will carry a system from a given initial state to the desired goal state. Many planners are domain-independent, allowing their deployment in a variety of domains. Such is the broad family of \gls{colin} planners. These planners perform Forward Search and call a \gls{MIP} solver multiple times at every state to check for consistency and to set bounds on the numeric variables. These checks can be computationally costly, especially in real-life applications. This paper suggests a method for identifying information about the specific state being evaluated, allowing the formulation of the equations to facilitate better solver selection and faster \gls{MIP} solving. The usefulness of the method is demonstrated in six domains and is shown to enhance performance significantly.

\end{abstract}

\glsresetall
\section{Introduction}

Automated Planning (often called AI Planning) is concerned with formulating a sequence of actions that transforms a system from a given initial state into a desired goal state. One strength of AI Planning is domain-independence: a single general planner can plan in a wide range of different application domains. Examples of domains in which Planning was used include space \cite{Chien2000aspen-}, battery usage \cite{Fox2011automatic}, and software penetration testing \cite{Obes2013attack}.   To facilitate application in realistic problems, planners need to reason with expressive models of the world.  Such models can be temporal: finding a plan with timestamped actions, taking into account action durations and concurrency, and numeric: considering variables that change discretely, or continuously, over time~\cite{fox2003pddl2}.

To solve expressive problems that contain temporal constraints, the planner requires a scheduler -- A technique for assigning values to the action's timestamps that would result in a valid plan. For example, the temporal planners SAPA~\cite{sapa} and Temporal Fast Downward~\cite{tfd} utilize the decision epoch mechanism, Crikey 3~\cite{crikeyaij} uses \gls{STN}. To solve hybrid problems that contain temporal as well as discrete and continuous change, the planner would require more complex approaches. For instance, SMTPlan formulating the problem into SAT  \cite{smtplan}, ENHSP using interval relaxation \cite{ENHSPscala2016interval}, DiNo discretizing time \cite{piotrowski2016heuristic}, and qtScoty using convex optimization \cite{fernandez2017mixed}. 

This work focuses on a family of planners that uses \gls{MIP} solvers to schedule the plan. This family includes COLIN \cite{coles2012colin}, POPF \cite{coles2010popf}, and OPTIC \cite{benton2012popf}. Planners from this family were used in a variety of real-world applications including robotics, \cite{popfinrobotics}, \gls{AUV} control \cite{popfauv}, \gls{MAV} control \cite{opticmav} and space applications \cite{opticspace,denenberg@2018automated}. These planners perform forward state-space search starting from the initial state. At each search state, an \gls{MIP} solver is used once to determine whether a consistent schedule for the plan exists. 
If no consistent schedule exists, the search branch can be pruned. If the state is consistent, the \gls{MIP} is then used several more times to bound the numeric variables and thus tightening the space of applicable actions in this state, narrowing the search space ahead. Recent work \cite{denenberg2019evaluating} has shown that solving LPs at every state multiple times may cause the search process to become slow and ineffective. 

The contribution of this work is twofold:
\begin{enumerate}
    \item We propose a better translation from the search state to \gls{MIP}. Our new compilation allows the better selection of optimization tools required for the consistency check of the problem, possibly eliminating the necessity of an \gls{MIP} solver.
    \item We propose methods for using information from a given state to compute the current variable bounds in the search, thus calling the solver fewer times and improve the performance of any planner from the \gls{colin} family.
\end{enumerate}

The rest of the article is ordered as follows: \secref{sec:background} describes the problem and the current way \gls{colin} planners solve it. \secref{sec:InfSelection} presents the suggested methodology. \secref{sec:Eval} presents the performance of the suggested improvements in a variety of different domains.

\section{Background}\label{sec:background}
\subsection{Problem Definition}\label{ssec:probdef}
A temporal planning problem with discrete and linear continuous numeric effects is a tuple:
\begin{equation}
\left\langle \gls{P},\gls{V}, \gls{I},\gls{G}\right\rangle
\end{equation} 

where \gls{P} is a set of propositions and \gls{V} a set of numeric variables. A state $S$ is defined as a set of value assignments to the variables in \gls{P} and \gls{V}. \gls{I} is such a set representing the initial state of the system. \gls{G} is the goal: a conjunction of propositions in $P$, and linear numeric conditions over the variables in $V$, of the form $w^1v^1 + w^2v^2 + ... + w^iv^i \{<,\leq,=,\geq\,>\} c$ ($w^1...w^i$ and $c \in  \mathds{R}$ are constants). An action changes the values and carries the system from one state to another. An action is defined as a tuple as well:
\begin{equation}
\left\langle\gls{dur},\gls{pres},\gls{effs},\gls{preinv},\gls{effinv},\gls{pree},\gls{effe}\right\rangle
\end{equation} 
where \gls{dur} is the duration of the action constrained by a conjunction of numeric conditions. \gls{pres} and \gls{pree} are conjunctions of preconditions (facts and numeric conditions) that must be true at the start and end of the action, \gls{preinv} are invariant conditions (preconditions that must hold throughout the action's duration), \gls{effs} and \gls{effe} are instantaneous effects that occur at the start and end of the action. Such effects may add or delete propositions $p\in\mathbf{P}$ (\gls{effadd}, \gls{effdelete}) or update a numeric variable $v^i \in V$ according to a linear instantaneous change:
\begin{equation}\label{eq:inst}
u \{\mbox{+=},=,\mbox{-=}\} w^1v^1 + w^2v^2 + ... + w^iv^i + c
\end{equation} 
where $u,v^i \in \gls{V}$ are numeric variables, and $c, w^i \in \mathds{R}$ are weights. \gls{effinv} is a conjunction of continuous effects that act upon numeric variables throughout the action's duration. In this work, we assume all change is linear and is of the form:
\begin{equation}\label{eq:cont}
\frac{\mathrm{d}v}{\mathrm{dt}} \left\{\mbox{+=},=,\mbox{-=}\right\} c
\end{equation} 
where $c \in \mathds{R}$ is a constant.

The planner is required to find a set of actions in \gls{A} and their schedule, that would carry the system from the initial state to the goal state.

\subsection{Running Example}\label{ssec:Exmp}
 
\begin{figure}[!tb]
	\centering
	\includegraphics[width=.6\columnwidth]{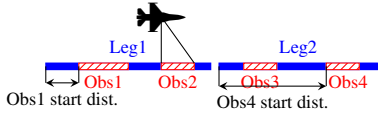}
	\caption{The flying observer}
	\label {fig:flyobs}
\end{figure}

\cite{denenberg2019evaluating} first introduced this example; it is an anonymized model of a real-life problem. In this domain, named flying observer, the planner is required to plan an \gls{UAV} observation mission. The \gls{UAV} is required to fly \emph{leg}s over a defined stretch of land containing objects to be observed. Each leg is of different length. Each observation has a different duration and requires a different type of equipment. A target-start distance defines the area within the leg in which the observation must take place.  The observation can only occur when the \gls{UAV} has flown more than the target-start distance of that leg ($\textnormal{flown}_l\geq\textnormal{target-start}_o$). A continuous numeric effect of the fly$_l$ action updates the distance flown so far in a leg: ${\frac{\mathrm{d}\textnormal{flown}_l}{\mathrm{dt}}} = \textnormal{Vel}_l$, where $\textnormal{Vel}_l$ is the flight velocity. {\tiny }

\figref{fig:flyobs} illustrates an instance of this domain: in this instance, two legs are defined (marked in solid blue lines). In each leg, two observations are required (marked in red, pattern-filled lines). All observations have a target-start distance defined, but for clarity, only the starting distance of the first and last observations are shown. 

In order to perform an observation, a defined piece of equipment needs to be calibrated and configured for a specific observation. Once the observation is done, the equipment needs to be released to become available for future observations. The domain comprises the following actions:

{\raggedright{
\noindent\textbf{take-off}$_l$(dur=5; \gls{pres}=\{on-ground,first-leg$_l$\}; \gls{effs}=\{$\neg\text{on-ground}$, flown$_l$=0\}; \gls{effe}=\{flying$_l$\}),

\noindent\textbf{set-course}$_{l1,l2}$(dur=1; \gls{pres}=\{done$_{l1}$, next$_{l1,l2}$\}; \gls{effs}=\{$\neg$done$_{l1}$\}; \gls{effe}=\{flying$_{l2}$, flown$_{l2}$=0\}),

\noindent\textbf{fly}$_l$(dur=\textsuperscript{distance$_l$}/\textsubscript{speed$_l$};  \gls{pres}=\{flying$_l$\};  \gls{preinv}=\{flown$_l$$\leq$distance$_l$\}; \gls{effe}=\{done$_{l}$, $\neg$flying$_l$\} \gls{effinv}=\{\textsuperscript{dflown$_l$}/\textsubscript{dt}+=1\}),

\noindent\textbf{configure}$_{o,e}$(dur=1; \gls{pres}=\{available$_e$, optionfor$_{o,e}$\}; \gls{effs}=\{$\neg$available$_e$\} \gls{effe}=\{configuredfor$_o$, pending$_{o,e}$\}),

\noindent\textbf{observe}$_{l,o}$(dur=\mbox{time-for$_o$}; \gls{pres}=\{configuredfor$_o$, contains$_{l,o}$, awaiting$_o$, \mbox{target-start$_o$}$\leq$flown$_l$\};  \gls{preinv}=\{flying$_l$\}; \gls{effs}=\{$\neg$awaiting$_o$\}; \gls{effe}=\{observed$_o$\}),

\noindent\textbf{release}$_{o,e}$ (dur=1; \gls{pres}=\{pending$_{o,e}$\};  \gls{effs}=\{$\neg$configuredfor$_o$, $\neg$pending$_{o,e}$\};  \gls{effe}=\{available$_e$\})

}}

\begin{figure}[tb]
	\centering		
	\includegraphics[width=.45\columnwidth]{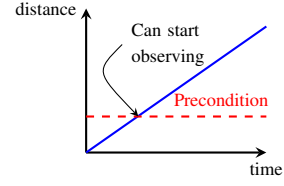}
	\caption{Distance Requirement} 
	\label{fig:dist}
\end{figure}

\begin{figure}[tb]
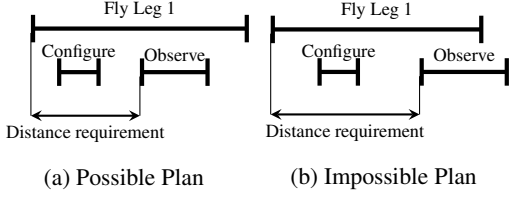

	\centering
	\begin{subfigure}{.4\columnwidth}
		\centering
		\includegraphics[width=1\linewidth]{Durative.tikz}
		\caption{Possible Plan}
		\label{fig:dur}
	\end{subfigure}	
	\begin{subfigure}{.4\columnwidth}
		\centering
		\includegraphics[width=1\linewidth]{impossible.tikz}
		\caption{Impossible Plan}
		\label{fig:imp}
	\end{subfigure}
	\caption{Durative Meaning of Distance Requirement}
	\label{fig:demo}
\end{figure}

The target distance precondition and temporal constraints force the observe actions to fit within the fly action. 
The meaning of the precondition is illustrated in \figref{fig:dist}: The blue line is a depiction of the distance change as the \gls{UAV} flies over the leg. The dashed red line is the precondition signifying the distance required for the start of the observation. When the distance reaches the value required in the precondition, the observation can start. Notice that this problem is, in fact, temporal, the numeric constraint can be easily converted to a temporal one depending on the manifestation in the temporal state can be seen in \figref{fig:demo}. 

\subsection{\gls{colin} and Forward Search} \label{ssec:OPTIC}

The \gls{colin} family of planners is based on the methodology of converting a state to an \gls{MIP} described in COLIN paper \cite{coles2012colin}. Here we survey that methodology.

To find a path from the initial state to the goal, \gls{colin} performs Forward Search. Starting from the initial state, \gls{colin} branches over applicable actions, exploring partially-ordered but un-time-stamped sequences of instantaneous actions. Durative actions are converted to a pair of instantaneous snap-action. Snap-actions mark the start ($A_\vdash$) and end ($A_\dashv$) of a durative action $A$. $A_\vdash$ has preconditions $\mathit{pre}_\vdash$A and effects $\mathit{eff}_\vdash$A; $A_\dashv$ is analogous. We define the set $\mathbf{A}_{inst}$ to contain all instantaneous actions in \gls{A}, including snap-actions.

A state $S$ in the search can be thought of as a set containing: propositions ($S.p \subseteq P$) that are true in $S$, and upper ($S.\mathit{max}(v)$) and lower ($S.\mathit{min}(v)$) bounds on the value each variable in $V$ can hold in $S$.  In the initial state, all variables have $\mathit{max}(v)=\mathit{min}(v)=v_I$ the value of $v$ specified in the initial state; $\mathit{max}(v)$ and $\mathit{min}(v)$ will only differ from each other in following states if a durative action with a continuous effect has acted on the variable. 

An action is deemed applicable if all its propositional invariants are satisfied by $S.p$ and if all numerical invariants can be satisfied by any value between $S.\mathit{max}(v)$ and $S.\mathit{min}(v)$. The planner compiles a list of all open applicable (named openlist). Search proceeds by popping the first state from the openlist: in our work, we use WA* (W=5), so sort the openlist by $h(S) + 5.g(s)$, using the temporal-numeric RPG heuristic of COLIN~\cite{coles2012colin}.

All successors $S'$ of $S$ are generated by adding or deleting all propositions in $\mathit{eff}^+_a$ and $\mathit{eff}^-_a$ respectively, and applying all discrete numeric effects to both $\mathit{max}(v)$ and $\mathit{min}(v)$ for all $v \in V$ affected by $\mathit{eff}^{num}_a$. This guarantees the $S'$ to be propositionally consistent. 

\gls{colin} then transforms all the temporal constraints to the following form:
\begin{equation}\label{eq:temporal}
Lb \leq t_j - t_i \leq Ub
\end{equation}
where $Lb,Ub \in \mathds{R}$ are the upper and lower bounds of a time interval. \gls{colin} also adds the necessary \emph{ordering} constraints to the plan. The action that has just been applied is ordered after the following: last actions to add each of its preconditions, actions whose preconditions it deletes, and actions with numeric effects on variables it updates or refers to in preconditions/effects. All ordering constraints are of the form {$t_j - t_i \geq \epsilon$}, where $t_j$,$t_i$ are the times at which the new and existing action must occur, respectively, and $\epsilon$ is a small positive constant. The temporal and ordering constraints, formulated as \eqref{eq:temporal}, constitute a \gls{STN} and the planner uses a \gls{STP} solver to check for temporal consistency. If the \gls{STP} can solve all equations (i.e., assign values to all time-steps such that the equations are valid), then the \gls{STP} was able to find a schedule, and the state is temporally consistent.

If $S'$ is propositionally and temporally consistent, the planner will compile the problem into an \gls{MIP} and then check for numerical consistency. Each step of the plan $i$ is given an \gls{MIP} variable $t_i$. Each variable $v \in \gls{V}$ is given three \gls{MIP} variables for each step $i$: $v_i$, $v'_i$, and $\delta v_i$. $v_i$ denotes the value of $v$ just before applying the action in step $i$, and $v'_i$ is the value right after the action's application. The planner applies \eqref{eq:inst} to the affected variable thus:
\begin{equation}
v'_i = v_i + w^1u^1_i + w^2u^2_i + ... + w^nu^n_i + c
\end{equation}
where $v,u\in\gls{V}$ are the numerical variables, $u^n_i$ is the value of the $n$th numerical variable at step $i$. $w^n$s are weights.

$\delta v_i$ is the value of the sum of all changes currently acting on $v$. Recall; in this work, each continuous effect is defined by a constant. When a new action is applied at step $prev$, the contribution of its effect is added to $\delta v_{prev}$; when an effect ends at step $i$, the value is removed from $\delta v_i$. 
\begin{equation}
\gls{dvi} = \left\{\begin{array}{lr}				   
\delta v_{i-1} + c_A & \qquad \mbox {if} A_i = A_\vdash\\
\delta v_{i-1} - c_A & \qquad \mbox{if} A_i = A_\dashv
\end{array}\right.
\end{equation}
The value of $v_i$ denoting the value just before the application of the action at step $i$ can be computed thus:
\begin{equation}\label{eq:numer}
	v_i = \gls{vprev} + \delta \gls{vprev} \left(t_i - \gls{tprev}\right)
\end{equation}
where $i$ is the current step index and $prev$ is the index of the last step in which the value of $v$ was computed. Note; if the calculation of the value of $v$ is required for an invariant in step $j$, the next time $v$ will be computed the time interval will be $t_i - t_j$, regardless whether $\delta v$ is changed or if the action is acting on $v$.

If the \gls{STN} or \gls{MIP} finds that the state $S'$ is inconsistent (i.e., there is no solution, no schedule that would enable achieving the state), it is pruned, and the search will not advance down that branch. If the state is consistent, then the \gls{MIP} solver is called two more times for each variable $v$ to optimize it and compute the new  $\mathit{max}(v)$ and $\mathit{min}(v)$ using the standard temporal/numeric relaxed planning graph heuristic of Colin~\cite{coles2012colin}. It is then inserted into the openlist, providing $h(S') \neq \infty$, i.e., the heuristic does not indicate $S'$ is a dead-end.

Note that when no numerical change is present, it is sufficient to use \gls{STN} to prove consistency. When continuous changes or numeric constraints are present, the \gls{MIP} solver is required for the proof of consistency.

In addition to the consistency check, the \gls{MIP} is also used to compute the bounds $\mathit{max}(v)$ and $\mathit{min}(v)$ of $S'$. If all continuous actions on $v$ have ended, then the last defined value $v_i$ can be maximized and minimized to compute the bounds. If $v$ had a continuous effect start but has not yet ended, another time variable is added to the \gls{MIP} denoted $t_{now}$, representing the latest timestamp. For each variable with an active effect, a variable $v_{now}$ is added. $t_{now}$ is ordered after all other time steps, and $v_{now}$ is calculated using \eqref{eq:numer}. The \gls{MIP} solver then minimizes and maximizes $v_{now}$ to find the possible bounds.

\cite{denenberg2019evaluating} showed that though it was previously thought that calling the \gls{MIP} solver is beneficial both for state consistency and for the variable bound update, in large real-life domains, the calls to the \gls{MIP} solver may slow the search down. The premise was that the \gls{MIP} problems solved are small, and therefore the call to the off-the-shelf solver would not be computationally expensive. However, it  was shown that in some real-life applications, this was not true: when the state contained many actions and multiple variables, the \gls{MIP} grew large and the solving of which became slow. 

\tabref{tab:mip_eq} demonstrates the process of converting a state into an \gls{MIP}. The table shows the \gls{MIP} equations for the partial-plan: take-off, fly$_{l0}^\vdash$, observe$_{o1,l0}^\vdash$, observe$_{o1,l0}^\dashv$ (for conciseness we assume that take-off is instantaneous  and no configure actions are required). 

The first action receives a single time variable $t_0$. The second action $Fly_{l0}^\vdash$, receives a time variable $t_1$, which is ordered after $t_0$, and the value of $flown$ is computed. The value before the fly action is the initial assignment, which is 0. Since there is no instantaneous effect on the action's start, the value just after the application of the action is the same. The invariants on the value are enforced just after the beginning of the fly action. The same process is repeated for the next action $Observe_{o0}^\vdash$: assigning a time variable for the action, calculating $\textit{flown}$ before and after the application of the action, and enforcing invariants. Note the ordering constraints formulated as temporal constraints in all actions. Also, the $flown$ variable is computed at each step, and its value depends on the previous step. 

\begin{table}[tb]
	\centering
	\resizebox{\columnwidth}{!}{
		\begin{tabular}{|l|l|c|l|l|}
			\hline
			Step               & Action                                    &            variables            & constraints                                                     & comment                          \\ \hline\hline
			0                  & TakeOff                                   &              $t_0$              & $\geq 0$                                                        &                                  \\ \hline\hline
			\multirow{4}{*}{1} & \multirow{4}{*}{Fly$_{l0}^\vdash$}        &              $t_1$              & $ - t_0 \geq \epsilon$                                          & Step1 afer Step0                 \\ \cline{3-5}
			                   &                                           &          $flown\_l0_1$          & $=0$                                                            & Initial Assignlemt               \\ \cline{3-5}
			                   &                                           & \multirow{2}{*}{$flown\_l0_1'$} & $flown\_l0_1$                                                   & Value after action               \\ \cline{4-5}
			                   &                                           &                                 & $ \leq distance\_l0$                                            & Invariant                        \\ \hline\hline
			\multirow{7}{*}{2} & \multirow{7}{*}{Observe$_{o1,l0}^\vdash$} &              $t_2$              & $ - t_1 \geq \epsilon$                                          & Step2 after Step1                \\ \cline{3-5}
			                   &                                           & \multirow{3}{*}{$flown\_l0_2$}  & $=flown\_l0_1' + 1*(t_2 - t_1)$                                 & Value before action              \\ \cline{4-5}
			                   &                                           &                                 & $ \geq target\mbox{-}start\_o1$                                 & Start precondition               \\ \cline{4-5}
			                   &                                           &                                 & $ \leq l0\_dist$                                                & Invariant                        \\ \cline{3-5}
			                   &                                           & \multirow{3}{*}{$flown\_l0_2'$} & $=flown\_l0_2$                                                  & Value after action               \\ \cline{4-5}
			                   &                                           &                                 & $ \geq Target\_dist\_o1$                                        & Start precondition               \\ \cline{4-5}
			                   &                                           &                                 & $ \leq l0\_length$                                              & Invariant                        \\ \hline\hline
			\multirow{6}{*}{3} & \multirow{6}{*}{Observe$_{o1,l0}^\dashv$} &     \multirow{2}{*}{$t_3$}      & $-t_2 \geq \epsilon$                                            & Step3 after Step2                \\ \cline{4-5}
			                   &                                           &                                 & $-t_2 \leq time-for_{o1}$                                       & Action duration                  \\ \cline{3-5}
			                   &                                           & \multirow{2}{*}{$flown\_l0_3$}  & $=flown\_l0_2' + 1*(t_3 - t_2)$                                 & Value before action              \\ \cline{4-5}
			                   &                                           &                                 & $ \leq distance\_l0$                                            & Invariant                        \\ \cline{3-5}
			                   &                                           & \multirow{2}{*}{$flown\_l0_3'$} & $=flown\_l0_3$                                                  & Value after action               \\ \cline{4-5}
			                   &                                           &                                 & $\leq l0\_length$                                               & Invariant                        \\ \hline\hline
			\multirow{2}{*}{4} & \multirow{2}{*}{now}                      &            $t_{now}$            & $-t_3 \geq \epsilon$,$-t_2 \geq \epsilon$,$-t_1 \geq \epsilon$, & After All Steps     \\ \cline{3-5}
			&                                           &        $flown\_l0_{now}$        &$=flown\_l0_3' + 1*(t_{now} - t_3)$                            & Value Now           \\ \hline
		\end{tabular}
	}
	\caption{LP Equations of a Partial Plan}\label{tab:mip_eq}
\end{table}

The next section will describe a method for changing \eqref{eq:numer} in a way that would allow calling the \gls{MIP} solver fewer times, and compile certain problems containing numerical constraints and change as \gls{STP}.

\section{Informed Selection of Solver for Consistency and Update}\label{sec:InfSelection}

The \gls{colin} methodology described in the previous chapter was developed to accommodate the general case in which hybrid planning is to be done, covering all possible state types. The planner uses the general representation both in consistency check and in the variable update. It was shown in \cite{denenberg2019evaluating} that the general approach could, at times, lead to slow solving. 

In this section, we propose two new methods for identifying two specific cases that frequently arise in the state in real-life problems. In such cases, the use of an LP solver can be made redundant, facilitating faster solving. In other cases, information from the current state may be injected into the problem definition to allow for faster solving. 

The first method examines the latest added action that carried the system from state $S$ to state $S'$. The second involves a conversion of specific numeric constraints and effects into \gls{STN} form. Finally, we describe how both these processes can facilitate a more effective update of variable bounds.

\subsection{Observing the Latest Action}\label{ssec:observing}

As stated previously, \gls{colin} solvers attempt to prove inconsistency with an \gls{STP} first. Then if continuous numeric effects and numeric constraints are present in the current state, the planner compiles the problem as an \gls{MIP}. \gls{colin} planners treat each state in the most general way: in the general case, every action may render the new state inconsistent. However, using knowledge about previous states, some instances in which the \gls{MIP} solver can be avoided may be found.

Consider the state TakeOff$_\vdash$, TakeOff$_\dashv$, Fly$^{l_{1}}_\vdash$, configure$^{o_1,e_2}_\vdash$: The partial plan contains continuous numeric effects on the variable $flown_{l_0}$. Therefore,  to prove this partial plan consistent, \gls{colin} requires an \gls{MIP} solver. However, since the planner is performing forward search, to reach this state, the planner must have been in a previous state, which it found consistent: TakeOff$_\vdash$, TakeOff$_\dashv$, Fly$^{l_{1}}_\vdash$. The configure action that is added does not require the value of $flown_{l_0}$, and its effect is propositional.

Assume the state $S$ is propositionally, temporally, and numerically consistent. The new state $S'$, which is reached from $S$ by addition of action $A\in\mathbf{A}_{inst}$  cannot be rendered numerically inconsistent if $A$ does not contain any numerical effects or constraints. Furthermore, since \gls{colin} only examines states $S'$ that are generated to be propositionally consistent, the state only has to be tested only for temporal consistency. 

Thus, if an added action $A$ contains only propositional or temporal constraints and effects, the state $S'$ can be deemed consistent by using the \gls{STP}, and an \gls{MIP} is not required. 

Note that if this test has determined that $S'$ is consistent, there is no need for a numerical variable bound update, as those do not change.

\subsection{Reformulation of \gls{MIP}}\label{ssec:reformulation}

During the search, \gls{colin} compiles the state into an \gls{MIP}, as depicted in the previous section. This transformation is done in a step-wise manner, meaning each step is transformed into a set of equations, and each step builds on the previous one. No consideration is taken as to what effect a step has on a variable; as long as the value of the variable is required, its value will be computed, and the next step will use said computed value. The notation \gls{tprev} denotes the previous step at which the value of variable $v$ was calculated, and the next step $i > prev$ that computes the value of $v$ will use the value stored in \gls{vprev}.

Here we suggest making a distinction between two types of steps that affect the value of $v$: A step containing a start or end of a continuous numeric effect and a step that does not. The later is a step containing numerical constraints on $v$ but does not change the value of $\delta v$. To distinguish between the two, we propose two notations $eff$ for steps that start or end an effect and $const$ for steps containing only constraints.  

Using the new notation \gls{tpe} would be the last time in which $v$ had an effect start or end, \gls{vpe} would be the value of $v$ calculated at that time point. Then \eqref{eq:numer} is written thus:
\begin{equation}\label{eq:newnumer}
		v_i = \gls{vpe} + \delta \gls{vpe} \left(t_i - \gls{tpe}\right)
\end{equation}
The conversion of the equations is demonstrated in the  previous example: this is presented in \tabref{tab:mip_eq_conv}. Notice the difference between \tabref{tab:mip_eq}: The effect acting on the variable $flown\_l0$ started in step 1, and therefore the computation of $flown\_l0_2$ $flown\_l0_3$ is always done with respect to step 1. 

\begin{table}[tb]
	\centering
	\resizebox{\columnwidth}{!}{
		\begin{tabular}{|l|l|c|l|l|}
			\hline
			Step               & Action                                    &            variables            & constraints                                                     & comment             \\ \hline\hline
			0                  & TakeOff                                   &              $t_0$              & $\geq 0$                                                        &                     \\ \hline\hline
			\multirow{4}{*}{1} & \multirow{4}{*}{Fly$_{l0}^\vdash$}        &              $t_1$              & $ - t_0 \geq \epsilon$                                          & Step1 afer Step0    \\ \cline{3-5}
			                   &                                           &          $flown\_l0_1$          & $=0$                                                            & Initial Assignlemt  \\ \cline{3-5}
			                   &                                           & \multirow{2}{*}{$flown\_l0_1'$} & $flown\_l0_1$                                                   & Value after action  \\ \cline{4-5}
			                   &                                           &                                 & $ \leq distance\_l0$                                            & Invariant           \\ \hline\hline
			\multirow{7}{*}{2} & \multirow{7}{*}{Observe$_{o1,l0}^\vdash$} &              $t_2$              & $ - t_1 \geq \epsilon$                                          & Step2 after Step1   \\ \cline{3-5}
			                   &                                           & \multirow{3}{*}{$flown\_l0_2$}  & $=flown\_l0_1' + 1*(t_2 - t_1)$                                 & Value before action \\ \cline{4-5}
			                   &                                           &                                 & $ \geq target\mbox{-}start\_o1$                                 & Start precondition  \\ \cline{4-5}
			                   &                                           &                                 & $ \leq l0\_dist$                                                & Invariant           \\ \cline{3-5}
			                   &                                           & \multirow{3}{*}{$flown\_l0_2'$} & $=flown\_l0_2$                                                  & Value after action  \\ \cline{4-5}
			                   &                                           &                                 & $ \geq Target\_dist\_o1$                                        & Start precondition  \\ \cline{4-5}
			                   &                                           &                                 & $ \leq l0\_length$                                              & Invariant           \\ \hline\hline
			\multirow{6}{*}{3} & \multirow{6}{*}{Observe$_{o1,l0}^\dashv$} &     \multirow{2}{*}{$t_3$}      & $-t_2 \geq \epsilon$                                            & Step3 after Step2   \\ \cline{4-5}
			                   &                                           &                                 & $-t_2 \leq time-for_{o1}$                                       & Action duration     \\ \cline{3-5}
			                   &                                           & \multirow{2}{*}{$flown\_l0_3$}  & $=flown\_l0_1' + 1*(t_3 - t_1)$                                 & Value before action \\ \cline{4-5}
			                   &                                           &                                 & $ \leq distance\_l0$                                            & Invariant           \\ \cline{3-5}
			                   &                                           & \multirow{2}{*}{$flown\_l0_3'$} & $=flown\_l0_3$                                                  & Value after action  \\ \cline{4-5}
			                   &                                           &                                 & $\leq l0\_length$                                               & Invariant           \\ \hline\hline
			\multirow{2}{*}{4} & \multirow{2}{*}{now}                      &            $t_{now}$            & $-t_3 \geq \epsilon$,$-t_2 \geq \epsilon$,$-t_1 \geq \epsilon$, & After All Steps     \\ \cline{3-5}
			                   &                                           &        $flown\_l0_{now}$        & $=flown\_l0_1' + 1*(t_{now} - t_1)$                             & Value Now           \\ \hline
		\end{tabular}
	}
	\caption{New LP Equations of a Partial Plan}\label{tab:mip_eq_conv}
\end{table}

If no continuous numeric actions have been acting on $v$ before the last action at \gls{tpe}, then the value of \gls{vpe} is a constant. This is seen in our example. The value of $flown_l0_1'$ is the same as the initial assignment. This conversion extremely useful when all constraints are of the form:
\begin{equation}\label{eq:simple_num_const}
	v \leq C
\end{equation}
where $v\in\gls{V}$ and $C \in \mathds{R}$. This constraint is written as a less than-equal-to constraint. Without loss of generality, this includes all constraints that have a single variable on one side and a constant on the other. If \gls{vpe} is constant, then using \eqref{eq:newnumer} when enforcing \eqref{eq:simple_num_const} at step $i$, we can write
\begin{equation}\label{eq:conv_numeric}
t_i - \gls{tpe} \leq \frac{\left(C- \gls{vpe} \right)}{\delta \gls{vpe} }
\end{equation}
Notice that at step $i$, all the variables on the right-hand side of \eqref{eq:conv_numeric} are known and are constant. Therefore, \eqref{eq:conv_numeric} can be formulated for each step $i$ as a constraint of the form of \eqref{eq:simple_num_const} as long as \gls{vpe} is constant. Notice that the numerical constraint in \eqref{eq:simple_num_const} is converted to a temporal constraint. If all constraints in the state can be converted thus, then the problem is, in fact, an \gls{STN}, and the \gls{MIP} solver is not required.

The example given above can be converted in such a way. The constraint $flown\_l0_2 > Target\_dist\_o1$ can be converted to $t_2 - t_1 > ^{\left(Target\_dist\_o1 - flown\_l0_1\right)} / _{1}$. All numeric constraints in this partial plan can be converted in the same way. This means that even though continuous numerical effects and numerical constraints are present, this problem is temporal, and can be solved with an \gls{STP}. 

If all numerical constraints were converted to temporal, the planner could determine consistency using the \gls{STN}. However, updating the bounds is still required. Observing \eqref{eq:numer}, we note that since \gls{vpe} is constant, the maximum and minimum of $v_i$ are dependent on the interval 
\begin{equation}\label{eq:interval}
T_i=\left(t_i - \gls{tpe}\right)
\end{equation}
The minimal size of the interval is zero. The maximum may be drawn from the state: if another constraint exists  such that limits $t_i$ or if \gls{tpe} is a start action beginning an effect, and is the only continuous numeric effect present, the maximal interval is the duration of the action. If no such value can be derived from the problem, then the interval is set to infinity. If $\delta v_i > 0$ then the minimum value for $v_i$ is when $T_i=0$ and is $\min \left(v_i\right)=\gls{vpe}$; and the maximum is $\max\left(v_i\right) = \gls{vpe} + \delta v_{eff} T_i$. The case in which $\delta v_i <0$ is analogous. 
This update method does not require an external solver and, therefore, very fast.

\subsection{Efficient Variable Update}\label{ssec:variable_update}

\cite{denenberg2019evaluating} have shown that the variable update is often the task that is most computationally expensive as it requires several calls to the \gls{MIP} solver, depending on the number of numerical variables in the partial plan. The previous section has detailed several cases in which the variable update can be avoided or done without the use of an \gls{MIP}. Here we attempt to facilitate faster \gls{MIP} solving in the update phase in case it is still required.

The \gls{MIP} solver may use one of several optimization methods; however, the selection of the method, as well as the method speed, depend on whether the feasible space is bounded and in which direction. We would like to supply the solver with information about variable boundlessness. We cannot use the bounds from the previous state $S$ as those might have changed by action $A$.

Therefore, we again examine the snap action $A$ added in the last step that carried the system from previous state $S$ to the current state $S'$. Recalling state $S$ contains bounds on $v$ ($S.\mathit{max}(v)$ and$S.\mathit{min}(v)$). We wish to determine whether the action $A$ is capable of expanding the limits of $v$, causing the interval $\left[\mathit{min}(v),\mathit{max}(v)\right]$ to grow if $A$ causes the bounds to contract, or if the interval retains its size but shifts.

If an instantaneous numeric effect exists, then the bounds on the latest defined $v_i$ (the step at which all continuous effects have ended, or $v_{now}$) can either be shifted due to an increase or decrease. An assignment would make the latest value a constant ($\mathit{max}(v)=\mathit{min}(v)=$Assigned value). If snap action $A$ contains a continuous numeric effect on $v$, then the bounds of $v$ may expand. Therefore, when solving the \gls{MIP} to update the bounds of $v$, the bounds of $v_{now}$ are defined as $\left[-\infty,\infty\right]$. If snap action $A$ does not contain a continuous numeric effect on $v$ or an instantaneous effect on $v$, then the bounds can only contract. Therefore, the bounds from $S$ are passed to the \gls{MIP} solver, leading to a smaller search space and faster update.

\section{Evaluation}\label{sec:Eval}

\begin{table}[!tb]
	\centering
	\resizebox{0.6\columnwidth}{!}{
		\begin{tabular}{|c|c|c|c|}
			\hline
			\multirow{2}{*}{Instance} & \multirow{2}{*}{Observations} & \multirow{2}{*}{Legs} & \multirow{2}{2.5cm}{Observations Required in Goal}\\ 
			& & & \\
			\hline
			1&10&28&4\\ \hline
			2&15&38&6\\ \hline
			3&20&48&8\\ \hline
			4&25&58&10\\ \hline
			5&30&68&12\\ \hline
			6&40&78&14\\ \hline
			7&40&88&16\\ \hline
			8&40&88&18\\ \hline
			$\vdots$&$\big\vert$&$\big\vert$&$\vdots$\\ \hline
			17&40&88&36\\ \hline
		\end{tabular}
	}
	\caption {Single Observation Per-Leg Instances}
	\label{tab:SingleInst}
\end{table}

\begin{table*}[tb]
	\centering
	\resizebox{\linewidth}{!}{
		\begin{tabular}{|l|c|c|c|c|c|c|c|c|c|c|c|c|c|c|c|c|c|c|c|c|c|c|c|}
			\hline
			\multicolumn{24}{|c|}{Flying Observer} \\ \hline
			 &                                             \multicolumn{15}{c|}{No limit on configure}                                              &                \multicolumn{8}{c|}{Configure only when flying}                \\ \hline
			    Instance              &  1   &  2   &   3   &   4    &   5    &   6    & 7 &   8   &   9   &  10   &  11   &  12   &   13   &   14   &   15   &  1   &   2   &   3    &   4    &   5    &   6    &  7    & 8   \\ \hline
			OPTIC                     & 0.52 & 6.58 & 90.15 & 172.82 & 268.59 & 808.65 & X & 33.15 & 55.62 & 66.42 & 82.5  & 91.19 & 456.68 & 831.02 &   X    & 6.99 & 35.96 & 105.62 & 213.33 & 237.86 & 538.46 &   X   & X\\ \hline
			Sec 3.3                   & 0.55 & 5.19 & 76.23 & 164.19 & 257.38 & 795.11 & X & 31.22 & 52.16 & 63.58 & 79.71 & 89.87 & 445.21 & 825.16 &   X    & 6.31 & 34.09 & 99.76  & 198.00 & 224.11 & 504.43 &   X   & X\\ \hline
			Sec 3.1                   & 0.19 & 2.91 & 37.11 & 85.15  & 149.76 & 578.61 & X & 9.32  & 13.63 & 19.84 & 26.14 & 26.59 & 192.26 & 441.77 & 924.58 & 1.69 &  7.91 & 27.62  & 58.88  & 68.15  & 442.05 & 332.87& X\\ \hline
			Sec 3.1+3.2               & 0.17 & 1.83 & 31.51 & 77.38  & 141.17 & 560.44 & X & 8.30  & 11.83 & 15.88 & 20.91 & 22.07 & 163.18 & 387.70 & 854.78 & 1.17 &  6.60 & 22.58  & 49.40  & 56.87  & 93.32  & 236.55& X\\ \hline
			Sec 3.1+3.3               & 0.21 & 2.31 & 35.51 & 84.23  & 149.12 & 580.07 & X & 9.34  & 13.70 & 19.81 & 22.89 & 26.53 & 190.98 & 434.36 & 924.04 & 1.64 &  7.81 & 27.35  & 57.95  & 65.76  & 441.35 & 336.51& X\\ \hline
			OPTIC-II                  & 0.15 & 1.88 & 32.91 & 79.63  & 140.55 & 557.89 & X & 8.29  & 11.07 & 14.91 & 21.82 & 24.50 & 167.22 & 391.78 & 857.92 & 1.23 &  5.98 & 20.40  & 49.15  & 56.91  & 91.02  & 241.99& X  \\ \hline
			\hhline{======================}
			\multicolumn{22}{|c|}{Factory Floor QA} \\ \cline{1-22}
			&                                  \multicolumn{13}{c|}{No limit on calibrate}                                  &                \multicolumn{8}{c|}{Calibrate only when manufacturing}                \\ \cline{1-22}
			Instance &  1   &  2   &   3   &   4    &   5    &   6    & 7 &   8   &   9   & 10&   11   &   12   & 13 & 1    &   2   &   3    &   4    &   5    &   6    &   7   & 8 \\ \cline{1-22}
			OPTIC    & 0.47 & 6.34 & 44.70 & 293.68 & 547.98 & 828.47 & X & 30.90 & 63.54 & X & 594.82 & 906.96 & X  & 6.35 & 36.64 & 122.41 & 267.91 & 314.17 & 452.16 & 880.83 & X \\ \cline{1-22}
			Sec 3.3  & 0.45 & 6.89 & 53.80 & 302.40 & 535.98 & 807.05 & X & 26.95 & 48.61 & X & 526.68 & 814.11 & X  & 5.83 & 34.54 & 118.00 & 254.39 & 294.76 & 424.92 & 827.24 & X \\ \cline{1-22}
			Sec 3.1  & 0.33 & 7.09 & 53.38 & 275.61 & 478.96 & 733.08 & X & 28.64 & 59.74 & X & 570.86 & 872.45 & X  & 5.36 & 30.56 &  92.71 & 205.55 & 232.52 & 351.25 & 726.83 & X\\ \cline{1-22}
		  Sec 3.1+3.2& 0.33 & 5.80 & 43.69 & 254.10 & 463.60 & 730.79 & X & 24.50 & 54.89 & X & 550.04 & 874.37 & X  & 5.18 & 29.23 &  90.89 & 201.99 & 234.87 & 347.51 & 725.29 & X\\ \cline{1-22}
		  Sec 3.1+3.3& 0.35 & 6.74 & 47.87 & 256.16 & 451.83 & 714.85 & X & 26.82 & 56.17 & X & 538.03 & 831.99 & X  & 4.45 & 24.65 &  85.27 & 192.05 & 219.80 & 325.14 & 690.79 & X\\ \cline{1-22}
			OPTIC-II & 0.33 & 6.92 & 47.19 & 253.12 & 465.30 & 722.89 & X & 26.65 & 56.25 & X & 548.36 & 843.43 & X  & 4.19 & 22.36 &  78.29 & 191.23 & 215.33 & 327.49 & 687.34 & X \\ \cline{17-22}
			\hhline{==================}
			\multicolumn{18}{|c|}{Single Rover}\\ \cline{1-18}
			\multicolumn{1}{|c|}{Instance} & 1    & 2    & 3    & 4    & 5    & 6    & 7    & 8     & 9    & 10     & 11   & 12   & 13     & 14   & 15   & 16    & 17\\ \cline{1-18}
			OPTIC                          & 0.08 & 0.04 & 0.09 & 0.18 & 0.35 & 0.88 & 7.80 & 12.03 & 0.29 & 374.94 & 0.22 & 0.06 & 225.89 & 2.98 & 0.66 & 36.23 & X \\ \cline{1-18}
			OPTIC-II                       & 0.07 & 0.04 & 0.10 & 0.18 & 0.35 & 0.96 & 8.13 & 13.09 & 0.23 & 375.69 & 0.23 & 0.06 & 221.19 & 3.04 & 0.65 & 36.68 & X \\ \cline{1-18}
			\hhline{===========}
			\multicolumn{11}{|c|}{Linear Generator}\\ \cline{1-11}
			\multicolumn{1}{|c|}{Tanks} & 10   & 20   & 30    & 40     & 50     & 60    & 70    & 80     & 90     & 100\\ \cline{1-11}
			OPTIC                       & 0.61 & 3.28 & 12.14 & 130.89 & 554.31 & 48.14 & 91.27 & 160.68 & 300.70 & X \\ \cline{1-11}
			OPTIC-II                    & 0.63 & 4.18 & 15.36 & 127.18 & 559.67 & 47.71 & 91.58 & 159.89 & 301.27 & X \\ \cline{1-11}
		\end{tabular}}
	\caption{Results}
	\label{tab:Res}
\end{table*}


In this section, we examine the performance of the proposed changes in six different domains, which stem from four physical world examples. The domains brought here were chosen to demonstrate both the strengths and the weaknesses of the contribution: The first four illustrate the family of cases in which the contribution is meaningful, shedding light not only to the benefits of our suggestion but also to ways in which to better model a problem for any \gls{colin} family planner. The other two domains are brought to demonstrate cases in which the contribution is not useful, in an attempt to assess the price of using it.  

As this work's contribution is in improving the OPTIC family of planners, we compare our suggestions with the performance of the latest implementation of OPTIC, and apply our improvements to the same code base, and name it OPTIC with Injected Information (OPTIC-II). 

The results and results of other additional tests on various IPC domains are publicly available at extra data\footnote{This is currently given in the extra data. When paper is accepted it will be published publicly online}.

All tests were performed on an Intel i7-8550U CPU$@$1.80GHz$\times$8 with 8GB RAM. The PDDL files of the domains, the respective problems, and result data will be published on a public website. Runtime results are presented in \tabref{tab:Res}. ``X'' in the table means the runtime was over 1000s. Since the domains and processes are deterministic no statistical analysis is required, as results may only vary due to CPU noise.

\subsection{Domains}

\subsubsection{Flying Observer}

We demonstrate \gls{opticii} on a set of previously published domains. The first is the flying observer that \cite{denenberg2019evaluating} has presented as a domain stemming from the industry. Two distinct models of this physical domain were tested: The first is identical to the running problem. The second variant of this domain has a requirement that the observer is flying before configuring or releasing pieces of equipment.

The problems in the first variant all require a single observation to be made in each leg. The instances are described in \tabref{tab:SingleInst}. 

In all instances of the second variant, six observations are defined for each leg, where the first and last leg both share an observation that cannot be performed in the first leg. The fact the observation cannot be used in the first leg may lead the planner down a branch, which would not be useful. The first instance contains two legs, the second three, and so on. \cite{denenberg2019evaluating} showed that these problems are challenging for the planner.

\subsubsection{Factory Floor \gls{QA}}

This domain describes \gls{QA} sampling planning on a factory floor. A machine produces parts at a specific rate; at some point, several parts are taken for sampling. The number of produced parts is limited for storage reasons. This domain is similar to the previous domain (the flown distance is analogous to produced parts); only here, we limit the total number of parts that can be produced, adding a global numerical invariant condition. In this domain, too, we have two variants - one allowing the calibration of measuring machinery before the beginning of the manufacturing process, and one that does not. As in the previous domain, the second variant is used for instances that require multiple samples of the same part. 

\subsubsection{Single Rover IPC Domain and Linear Generator}

The Single Rover domain is a standard domain taken from the standard IPC 3, and \cite{coles2012colin} used it to demonstrate the hybrid planning mechanism. 

The Linear Generator  is yet another standard domain that was widely used in previous papers. It describes a generator consuming fuel to generate energy. The generator may be refueled from auxiliary tanks. All actions affect the main-tank fuel quantity, and the fact that all actions contain numerical change and constraints it was expected \gls{opticii} to show little to no improvement in solving problems from this domain. 

\subsection{Results}

Implementing the changes suggested in this work requires additional tests before building an \gls{MIP}. These tests and checks come with a computational price. However, as can be seen in Table 4, that price is not high. In the Linear generator domain and the single rover, none of the changes are useful. In most states, the information from the applied action cannot reduce the computations, and the problem is not convertible to temporal.  This is because in many states, for instance,  there are often two continuous linear actions operating on the same variable. The changes not being useful mean that all checks will be false when running \gls{opticii}, and the \gls{MIP} solver will be used just as in OPTIC. The results show that \gls{opticii} indeed performs a bit slower when used in these domains.

In the four domains from the Flying Observer and QA, \gls{opticii} performs far better than OPTIC. We present the results of all three suggested changes, the contribution of each change separately, and possible combinations to better understand the results.

\subsubsection{Efficient Variable Update}

In \secref{ssec:variable_update}, we suggest using information about the current action to update the variables' bounds efficiently. This improves the \gls{MIP} solving in the variable update stage of the search. Therefore, we expect this change to be more prominent when the planner handles large plans that result in larger \gls{MIP} problems, and when a large amount of variables needs to be updated. 

In \tabref{tab:Res}, the line named ``Sec. 3.3'' presents OPTIC performance when only this change is present in the four domains taken from \cite{denenberg2019evaluating}. It can be seen that this change contributes to the performance is more prominent in higher instances where the plan is, indeed, quite long and contains many variables.

Though this change's contribution is not visible in simple academic domains, it helps scale large problems such that arise in a real-life domain and, therefore, useful.

\subsubsection{Observing the Latest Action}
In \secref{ssec:observing}, we suggested using information about the current action to decide whether an \gls{STP} can be used to prove temporal consistency even though a numeric change is present in  the state. This change may lower the number of \glsplural{MIP}  that will be solved during the search and, therefore, speed up the search. The runtime results of OPTIC running only this improvement is dubbed ``Sec. 3.1'' in the table.

In the first four domains, we see a significant reduction of \glsplural{MIP} solved in the search\footnote{The number of states proved consistent with an \gls{MIP} out of the visited states is given in the extra data, and will be published online}, which results in a faster search. In the last two domains, the reduction is minimal, and so the tests slightly slow the search process.

\subsubsection{Reformulating the \gls{MIP}}

The last contribution we examine is the one described in \secref{ssec:reformulation}. For implementation reasons, this change was only applicable to the previous change. This change was useful in both the flying observer domains. When applied to these domains, all problems were converted to temporal, and no \glsplural{MIP} were solved. The result was a significant improvement in planning speed.

In the \gls{QA} domains, only some of the states were converted from numerical to temporal, and therefore the change was less prominent. The planner was able to identify that the requirement for the total amount of produced parts was the variable that prevented the conversion. If the domain expert believes the total amount limit cannot be reached, he can remove that constraint from the domain and allow for much faster Planning. Thus, using this method, we can improve performance, and perform knowledge engineering, presenting the model expert with possible ways to improve the planning process.

\section{Conclusions}

This work presented three methods for improving the search in the OPTIC family of planners: injecting state information into the consistency check, injecting state information into the variables bound update, and reformulating the \gls{MIP} as an \gls{STN}. These suggested changes to the planner can improve \gls{MIP}'s solving time or, at times, help avoid using an \gls{MIP} solver altogether. These changes were shown to be relatively cheap and useful in many cases. These changes apply to a board and a popular family of planners.

Future work would include additional tests and profiling. Also, exploiting state information in other forward search planners can be examined. 

%

\bibliographystyle{aaai21}
\bibliography{aaai}

\end{document}